\documentclass[conference]{IEEEtran}
\IEEEoverridecommandlockouts

\usepackage{cite}
\usepackage{amsmath,amssymb,amsfonts}
\usepackage{algorithmic}
\usepackage{graphicx}
\usepackage{textcomp}
\usepackage{xcolor}
\usepackage{booktabs}
\usepackage{multirow}
\usepackage{url}
\usepackage[hidelinks,hypertexnames=false]{hyperref}
\usepackage{float}
\usepackage{afterpage}

\begin{document}

\title{3D MRI-Based Alzheimer’s Disease Classification Using Multi-Modal 3D CNN with Leakage-Aware Subject-Level Evaluation}

%Multi-Modal 3D CNN with Subject-Level Evaluation for Alzheimer’s %Disease Classification on Raw OASIS-1

\author{
    \IEEEauthorblockN{
        Md Sifat$^*$, 
        Sania Akter$^*$, 
        Akif Islam$^*$, 
        Md. Ekramul Hamid$^*$, \\
        Abu Saleh Musa Miah$^\dagger$, 
        Najmul Hassan$^\dagger$, 
        Md Abdur Rahim$^\ddagger$, 
        Jungpil Shin$^\dagger$
    }
    \IEEEauthorblockA{
        \small
        $^*$Department of Computer Science and Engineering, University of Rajshahi, Rajshahi, Bangladesh\\
        $^\dagger$School of Computer Science and Engineering, University of Aizu, Aizuwakamatsu, Japan\\
        $^\ddagger$Department of Computer Science and Engineering, Pabna University of Science and Technology, Pabna, Bangladesh\\
        Emails: \{mdsifat371, saniaaker, iamakifislam\}@gmail.com, ekram\_hamid@ru.ac.bd, \\
        \{musa, d8225102, jpshin\}@u-aizu.ac.jp, rahim@pust.ac.bd
    }
}

\maketitle

\begin{abstract}
Deep learning has become an important tool for Alzheimer's disease (AD) classification from structural MRI. Many existing studies analyze individual 2D slices extracted from MRI volumes, while clinical neuroimaging practice typically relies on the full three-dimensional structure of the brain. From this perspective, volumetric analysis may better capture spatial relationships among brain regions that are relevant to disease progression. Motivated by this idea, this work proposes a multi-modal 3D convolutional neural network for AD classification using raw OASIS-1 MRI volumes. The model combines structural T1 information with gray matter, white matter, and cerebrospinal fluid probability maps obtained through FSL FAST segmentation in order to capture complementary neuroanatomical information. The proposed approach is evaluated on the clinically labelled OASIS-1 cohort using 5-fold subject-level cross-validation, achieving a mean accuracy of $72.34\% \pm 4.66\%$ and a ROC-AUC of $0.7781 \pm 0.0365$. GradCAM visualizations further indicate that the model focuses on anatomically meaningful regions, including the medial temporal lobe and ventricular areas that are known to be associated with Alzheimer's-related structural changes. To better understand how data representation and evaluation strategies may influence reported performance, additional diagnostic experiments were conducted on a slice-based version of the dataset under both slice-level and subject-level protocols. These observations help provide context for the volumetric results. Overall, the proposed multi-modal 3D framework establishes a reproducible subject-level benchmark and highlights the potential benefits of volumetric MRI analysis for Alzheimer's disease classification.
\end{abstract}


\section{Introduction}

Alzheimer's disease (AD) is the most common form of dementia and a major global health challenge, affecting more than 55 million people worldwide \cite{who2023}. Because AD is progressive and irreversible, earlier and more reliable diagnosis remains important for clinical monitoring, treatment planning, and supportive care. Structural magnetic resonance imaging (MRI) is therefore widely used in AD research because it can reveal neuroanatomical changes associated with disease progression, including hippocampal atrophy, cortical thinning, and ventricular enlargement \cite{jack2018nia}.

Deep learning has become a popular approach for automated AD classification from MRI, with many studies reporting very high accuracies on public datasets such as OASIS-1 \cite{marcus2007open}. However, high numerical performance does not necessarily imply meaningful clinical learning. A central question therefore arises: are these models truly learning disease-relevant neuroanatomical patterns, or are they relying on surrounding image characteristics, preprocessing artefacts, and subject-specific cues? This question is especially important in medical imaging, where strong predictive performance alone does not guarantee clinically meaningful learning.

One important factor influencing reported performance is how MRI data are partitioned. Because a volumetric MRI contains many highly similar 2D slices from the same subject, slice-level splitting can place related slices across training, validation, and test sets. In such patient-dependent settings, a model may appear highly accurate while failing to generalize at the subject level \cite{leakage2021, leakage2023}. A second concern involves dataset construction. Publicly redistributed versions of OASIS-1, especially Kaggle-derived slice datasets, are convenient to use but do not always clearly document subject selection, preprocessing, or label handling. In some cases, unlabeled subjects may effectively be treated as healthy controls, introducing further uncertainty into reported results.

Motivated by these concerns, this work proposes a leakage-aware multi-modal 3D CNN for Alzheimer's disease classification using raw OASIS-1 MRI volumes. The model integrates structural T1 information with tissue probability maps derived from segmentation to capture complementary neuroanatomical signals. To better understand discrepancies reported in the literature, we additionally conduct diagnostic experiments using slice-based datasets under different evaluation protocols.

The main contributions of this paper are as follows:

\begin{enumerate}
  \item We propose a multi-modal 3D CNN for Alzheimer's disease classification from raw OASIS-1 MRI volumes, integrating T1 intensity with GM, WM, and CSF probability maps.
  \item We establish a leakage-aware subject-level evaluation protocol using 5-fold stratified cross-validation on the clinically labelled OASIS-1 cohort.
  \item We provide interpretability analysis using GradCAM to examine whether model decisions align with known Alzheimer's-related anatomical regions.
  \item We report supporting diagnostic observations on a slice-based Kaggle-derived dataset to illustrate how evaluation protocols can affect reported performance.
\end{enumerate}

Rather than claiming clinical readiness, this study aims to provide a reproducible and leakage-aware subject-level benchmark for volumetric MRI-based AD classification.

\section{Related Work}
\label{sec:related}

Deep learning has become a widely used approach for Alzheimer's disease (AD)
classification from structural MRI. Recent studies have explored CNN,
transformer, and hybrid architectures on OASIS-derived MRI data. For example,
Almalki et al.~\cite{almalki2025} combined ResNet101 with a Vision Transformer
for four-class AD staging, while Ahmed~\cite{pseudocolor2025} used
pseudo-color MRI representations with a Vision Transformer. Dag et
al.~\cite{adnet2025} proposed a lightweight CNN for sagittal MRI slices, and
Ullah et al.~\cite{adaptivefusion2025} used adaptive fusion of ResNet50 and
ViT features. Other studies have also explored alternative representations,
such as topological feature fusion with DenseNet121~\cite{tda2026}. Although
these studies report very high accuracies, their results should be interpreted
cautiously because they differ in data representation, subject inclusion,
preprocessing, and train--test partitioning.

Many existing approaches operate on two-dimensional slices
extracted from volumetric MRI scans. Although slice-based
representations are computationally efficient, they can make
evaluation sensitive to the data-splitting protocol. In particular,
when highly similar slices from the same subject are divided
across training and testing sets, the reported accuracy may be
inflated compared with protocols that evaluate generalization
at the subject level.

This observation has been discussed in prior neuroimaging studies. Tufail et al.~\cite{tufail2020} evaluated several deep architectures on OASIS-1 using subject-level cross-validation and reported accuracies between 63\% and 65\%. Similarly, Yagis et al.~\cite{yagis2021} compared multiple CNN models and observed a decrease in accuracy when transitioning from slice-level to subject-level evaluation protocols. Other analyses have also emphasized the importance of carefully designing data partitioning strategies for volumetric medical imaging datasets~\cite{leakage2021, leakage2023}. These works suggest that considering the subject as the primary unit of evaluation may provide a more clinically meaningful estimate of model generalization in MRI-based studies.

Explainable artificial intelligence (XAI) is also important in MRI-based
AD classification because high accuracy alone does not ensure clinically
meaningful learning. GradCAM~\cite{selvaraju2017gradcam} is commonly used to
examine whether model attention falls on AD-relevant regions rather than
background or preprocessing artefacts. In structural MRI, such attention is
expected around regions affected by neurodegeneration, including the
hippocampus, medial temporal lobe, entorhinal cortex, and ventricular
areas~\cite{jack2018nia}.

\subsection{Research Gap}

Despite strong reported results, three methodological gaps remain. First,
many existing methods rely on 2D slice-based representations, which may not
fully preserve volumetric neuroanatomical context. Second, reported
accuracies are difficult to compare across studies because evaluation
protocols differ, and slice-level splitting can introduce subject-level
leakage. Third, relatively few studies examine whether model decisions are
supported by AD-relevant anatomical regions. To address these gaps, this work
uses multi-modal 3D MRI inputs, strict subject-level cross-validation, and
GradCAM-based anatomical plausibility analysis.
% ============================================================

% ============================================================
\section{Methodology}
\label{sec:method}
This section presents the proposed leakage-aware multi-modal 3D CNN for Alzheimer's disease classification from raw OASIS-1 MRI volumes. The method is designed to preserve volumetric neuroanatomical structure, integrate complementary tissue-based information, and evaluate performance under strict subject-level validation.

\subsection{Dataset Description}

\subsubsection{Raw OASIS-1 Overview}

The Open Access Series of Imaging Studies (OASIS-1) \cite{marcus2007open} is a widely used public neuroimaging dataset containing cross-sectional T1-weighted structural MRI scans from 416 subjects aged 18--96 years. Among these, 100 subjects over the age of 60 were clinically diagnosed with very mild to moderate Alzheimer's disease. Dementia severity is annotated using the Clinical Dementia Rating (CDR) scale, where CDR values of 0, 0.5, 1.0, and 2.0 correspond to non-demented, very mild, mild, and moderate dementia, respectively.

\subsubsection{Labelled Subset Used in This Study}

Although OASIS-1 contains 416 subjects with 436 MRI scans, only 235 subjects have available CDR annotations suitable for supervised dementia classification. The remaining subjects do not include CDR labels and were therefore excluded from the experiments. The resulting clinically labelled cohort used in this study is summarized in Table~\ref{tab:oasis_dist} and forms the basis of the raw-volume experiments.

\begin{table}[!htbp]
\centering
\caption{Distribution of the Clinically Labelled OASIS-1 Cohort}
\resizebox{\columnwidth}{!}{%
\begin{tabular}{llccccc}
\toprule
\textbf{CDR} & \textbf{Label} & \textbf{N} & \textbf{Mean Age} & \textbf{Mean MMSE} & \textbf{Male} & \textbf{Female} \\
\midrule
0.0 & Non-demented  & 135 & 69.07 & 29.10 & 38 & 97 \\
0.5 & Very mild AD  &  70 & 76.21 & 25.64 & 31 & 39 \\
1.0 & Mild AD       &  28 & 77.75 & 21.68 &  9 & 19 \\
2.0 & Moderate AD   &   2 & 82.00 & 15.00 &  1 &  1 \\
\bottomrule
\end{tabular}
}
\label{tab:oasis_dist}
\end{table}

\begin{figure*}[!t]
\centering
\includegraphics[width=0.8\textwidth]{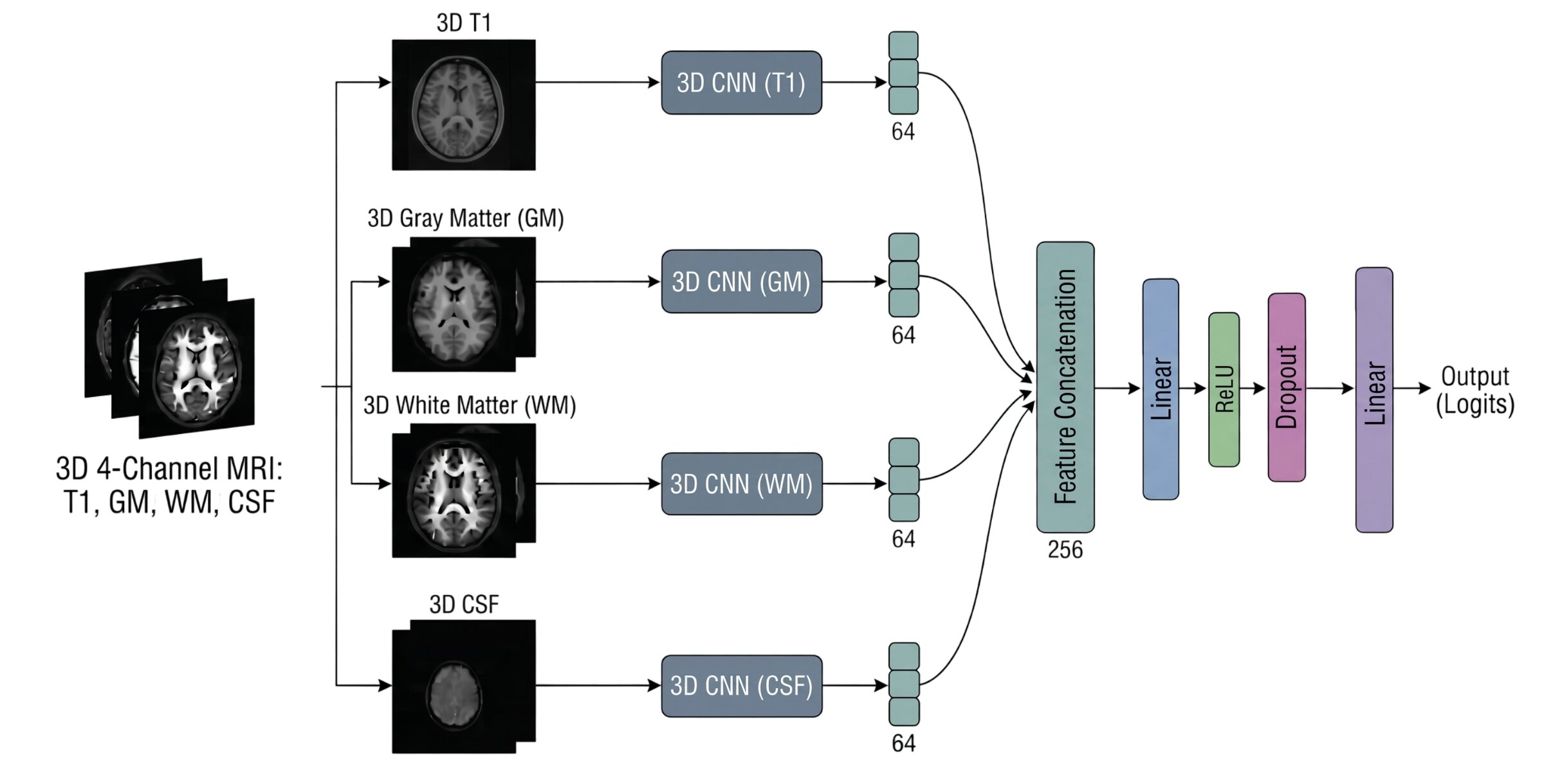}
\caption{Overview of the proposed multi-modal 3D CNN architecture, where modality-specific encoders process T1, GM, WM, and CSF inputs independently and their learned representations are fused through a shared late-fusion classification head for binary dementia classification.}
\label{fig:arch}
\end{figure*}

\subsubsection{Kaggle-Derived OASIS-1 Slice Dataset}

A widely used Kaggle redistribution of OASIS-1 provides the data as pre-extracted 2D JPEG slices instead of the original 3D MRI volumes. To examine its relationship with the raw dataset, subject identifiers were traced from slice captions and compared with the clinically labelled OASIS-1 cohort used in this study. The class-wise composition of the Kaggle-derived dataset is summarized in Table~\ref{tab:kaggle_dist}.

\begin{table}[!htbp]
\centering
\caption{Distribution of the Kaggle-Derived OASIS-1 Slice Dataset}
\begin{tabular}{lcc}
\toprule
\textbf{Class} & \textbf{Unique Subjects} & \textbf{Total Slices} \\
\midrule
Non-demented       & 285 & 67,222 \\
Very mild dementia &  58 & 13,725 \\
Mild dementia      &  21 &  5,002 \\
Moderate dementia  &   2 &    488 \\
\midrule
\textbf{Total}     & \textbf{366} & \textbf{86,437} \\
\bottomrule
\end{tabular}
\label{tab:kaggle_dist}
\end{table}

The Kaggle dataset contains 366 unique subjects, whereas the clinically labelled raw OASIS-1 subset used in this study includes 235 subjects. This difference indicates that the redistributed slice dataset does not correspond exactly to the labelled raw cohort and may reflect differences in subject inclusion or preprocessing procedures. Visual inspection of the slices also suggests the presence of residual skull boundaries and background regions that may influence 2D model behavior.

\subsection{Problem Formulation}
\label{sec:problem_formulation}

The clinically labelled OASIS-1 subset used in this study contains 235 subjects with Clinical Dementia Rating (CDR) annotations. Because the Moderate Dementia category contains only two subjects, fine-grained multi-class subject-level evaluation is not statistically stable. Therefore, the task is formulated as a binary classification problem in which subjects with CDR $= 0$ are assigned to the \textit{Non-Demented} class and all subjects with CDR $> 0$ are grouped into a single \textit{Demented} class. This yields 135 Non-Demented and 100 Demented subjects.

\subsection{Raw MRI Preprocessing Pipeline}
\label{sec:preprocess}

The proposed method operates directly on raw T1-weighted NIfTI volumes from the OASIS-1 dataset \cite{marcus2007open}. Each MRI volume originally has a spatial size of $256 \times 256 \times 128$. To generate anatomically aligned multi-modal inputs, preprocessing was performed using ANTsPy 0.6.1 and FSL.

N4 bias-field correction was first applied to reduce intensity non-uniformity \cite{tustison2010n4itk}. Non-brain tissues were then removed using FSL BET \cite{smith2002fsl}, followed by affine registration to the MNI152 template using FSL FLIRT for spatial alignment across subjects. Tissue segmentation was subsequently performed with FSL FAST \cite{zhang2001} to obtain gray matter (GM), white matter (WM), and cerebrospinal fluid (CSF) probability maps.

\begin{figure}[H]
\centering
\includegraphics[width=\columnwidth]{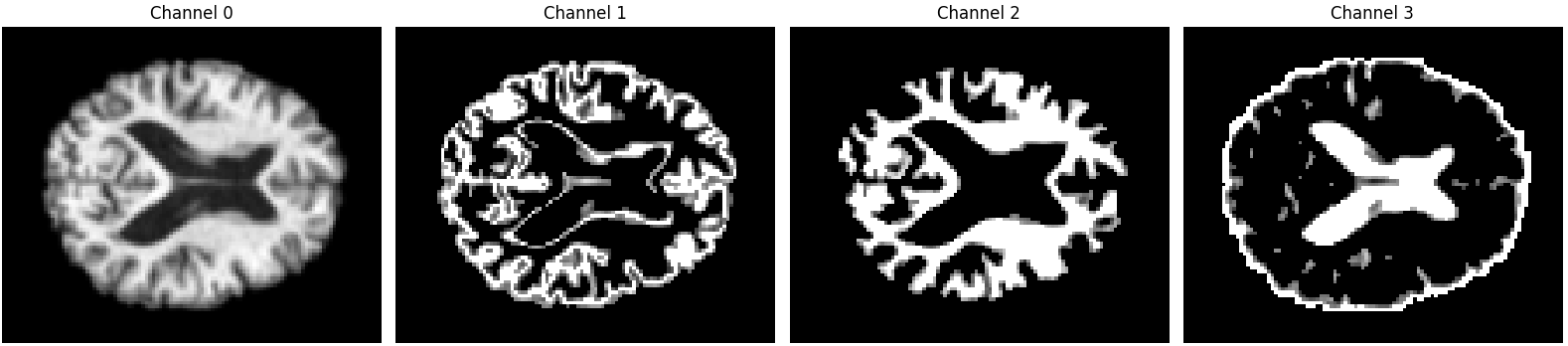}
\caption{Preprocessed multi-channel MRI input. Channel 0: bias-corrected T1 image; Channels 1--3: GM, WM, and CSF probability maps from FSL FAST segmentation.}
\label{fig:modal_channels}
\end{figure}

Finally, the bias-corrected T1 image and the three tissue probability maps were normalized and stacked into a four-channel representation. After preprocessing, each subject is represented as a 3D tensor of size $(91 \times 109 \times 91 \times 4)$ corresponding to T1, GM, WM, and CSF, as illustrated in Fig.~\ref{fig:modal_channels}.

\subsection{Supporting Diagnostic Experiments on the Slice-Based Dataset}
\label{sec:slice_protocol}

To provide additional context for the raw-volume results and to examine how evaluation protocols affect reported performance, two supporting diagnostic experiments were conducted on the Kaggle-derived OASIS-1 slice dataset. In the first experiment, slices were split at the slice level, replicating a common setting in the literature. In the second experiment, strict subject-level separation was enforced using a 70/15/15 subject-level train, validation, and test split, ensuring that no subject appeared in more than one partition.

\subsection{Proposed Multi-Modal 3D CNN}
\label{sec:architecture}

The proposed model is a multi-modal 3D convolutional neural network that takes four complementary inputs from each MRI volume: the bias-corrected T1 image and the corresponding gray matter (GM), white matter (WM), and cerebrospinal fluid (CSF) probability maps. As shown in Fig.~\ref{fig:arch}, the network follows a late-fusion design in which each modality is processed independently before the learned features are combined for final classification.

This design is motivated by the different roles of the input channels. The T1 volume preserves structural intensity information, while the GM, WM, and CSF maps provide tissue-specific cues that may reflect Alzheimer's-related changes more explicitly. In particular, gray matter loss and ventricular CSF enlargement are well-known structural markers of disease progression~\cite{jack2018nia}. Processing each modality separately therefore allows the network to learn features that are better matched to the characteristics of each signal before fusion~\cite{baltrusaitis2019, kong2022, multimodal_vit2024}.

Each modality-specific branch is implemented as a lightweight 3D CNN with three Conv3D--BatchNorm--ReLU blocks having output channels of 16, 32, and 64, respectively~\cite{dhinagar2021, peerj2025}. Max-pooling is applied after the first two blocks, and a final global average pooling layer produces a compact 64-dimensional embedding for each modality~\cite{lin2014gap}. The four embeddings are then concatenated into a 256-dimensional fused representation, which is passed through a shared classification head consisting of Linear$(256 \rightarrow 128)$, ReLU, Dropout$(p=0.30)$, and Linear$(128 \rightarrow 2)$ layers. Dropout is used to improve regularization and reduce co-adaptation among fused features~\cite{srivastava2014dropout}.

Overall, the proposed architecture learns modality-specific structural patterns and their joint interactions. The main contribution of this work lies not only in the lightweight multi-modal 3D architecture, but also in the leakage-aware subject-level evaluation and diagnostic comparison with slice-based protocols.

\subsection{Training and Leakage-Aware Evaluation Protocol}
\label{sec:evaluation_protocol}

The proposed model was evaluated on all 235 clinically labelled subjects using 5-fold stratified cross-validation. This protocol was designed to preserve the class distribution of the dataset, maintaining an approximately 57:43 ratio between Non-demented and Demented subjects in each fold. In each iteration, one fold, containing about 47 subjects, was held out for testing, while the remaining subjects were further divided into training and validation subsets using a 90/10 split.

A new model was trained independently for every fold so that each subject appeared in the test set exactly once across the full evaluation. To ensure strict subject-level separation, a post-hoc intersection check was also performed, confirming that no subject appeared simultaneously in the training, validation, and test partitions of any fold. This protocol was adopted to provide a more reliable estimate of subject-level generalization.

\subsection{Interpretability Analysis}
\label{sec:gradcam_method}

To examine whether the proposed model relied on anatomically meaningful patterns, GradCAM \cite{selvaraju2017gradcam} was applied to the trained 3D network. The resulting saliency maps were analyzed qualitatively to assess whether model attention was concentrated in brain regions commonly associated with Alzheimer's disease, such as the medial temporal lobe and ventricular areas, rather than irrelevant borders or background structures.

\section{Results and Discussion}
\label{sec:results}

\subsection{Results of the Proposed Multi-Modal 3D CNN on Raw OASIS-1}

The proposed multi-modal 3D CNN achieved a mean accuracy of $72.34\% \pm 4.66\%$ and a mean ROC-AUC of $0.7781 \pm 0.0365$ under 5-fold subject-level cross-validation on raw OASIS-1 MRI volumes. These results provide a reproducible subject-level benchmark for Alzheimer's disease classification using volumetric MRI and multi-modal structural inputs.

\begin{table}[!htbp]
\centering
\caption{Per-Fold and Mean Performance of the Proposed Multi-Modal 3D CNN}
\begin{tabular}{lcc}
\toprule
\textbf{Fold} & \textbf{Accuracy (\%)} & \textbf{ROC-AUC} \\
\midrule
Fold 1 & 76.6 & 0.7759 \\
Fold 2 & 72.3 & 0.7926 \\
Fold 3 & 63.8 & 0.7093 \\
Fold 4 & 76.6 & 0.8130 \\
Fold 5 & 72.3 & 0.8000 \\
\midrule
\textbf{Mean} & \textbf{72.34 $\pm$ 4.66} & \textbf{0.7781 $\pm$ 0.0365} \\
\bottomrule
\end{tabular}
\label{tab:m3_results}
\end{table}

As shown in Table~\ref{tab:m3_results}, performance remains reasonably stable across folds, with the strongest results observed in Folds~1 and~4 and the weakest in Fold~3. The fold-wise results suggest that the model identifies Non-demented subjects more reliably than Demented subjects, indicating that missed positive cases remain an important limitation in the current binary screening setting.

\begin{figure}[H]
\centering
\begin{tabular}{cc}
\includegraphics[width=0.44\columnwidth,height=3cm]{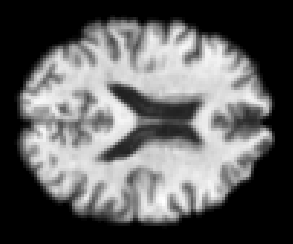} &
\includegraphics[width=0.44\columnwidth,height=3cm]{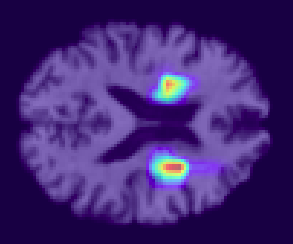} \\
{\scriptsize (a) Input MRI volume} & {\scriptsize (b) GradCAM saliency map}
\end{tabular}
\caption{Qualitative comparison between an input MRI volume and its corresponding GradCAM visualization in the proposed multi-modal 3D framework.}
\label{fig:m3_ori_grad}
\end{figure}

\begin{figure*}[!t]
\centering
\includegraphics[width=\textwidth]{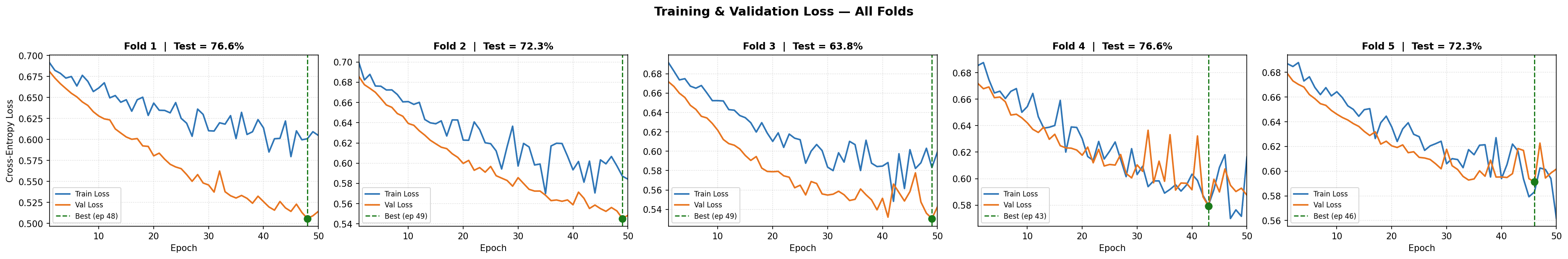}
\caption{Training and validation loss curves across the five subject-level cross-validation folds of the proposed multi-modal 3D CNN. The dashed vertical line in each panel marks the epoch of the selected best checkpoint for that fold.}
\label{fig:training_curves}
\end{figure*}

Most importantly, the GradCAM visualization in Fig.~\ref{fig:m3_ori_grad} shows activation concentrated around medial temporal and ventricular regions that are more consistent with known Alzheimer's-related structural changes. This suggests that the proposed model captures anatomically meaningful patterns that are more clearly localized in relevant brain regions than those observed in the supporting slice-based experiments. The training dynamics were also stable across folds, as illustrated in Fig.~\ref{fig:training_curves}.

\subsection{Supporting Diagnostic Experiments on the Slice-Based Dataset}

To provide additional context for the final raw-volume results, we also conducted two supporting diagnostic experiments on the Kaggle-derived slice dataset.

\subsubsection{Slice-Level Evaluation}

Under the original slice-level evaluation setting, the replicated slice-based pipeline achieved 99.19\% test accuracy. While this result appears highly competitive, it was obtained in a setting where highly similar slices from the same subject may appear across training and testing partitions.
\begin{figure}[!t]
\centering
\includegraphics[width=0.42\columnwidth,height=3cm]{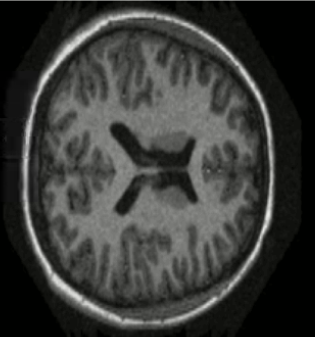}
\hspace{0.02\columnwidth}
\includegraphics[width=0.42\columnwidth,height=3cm]{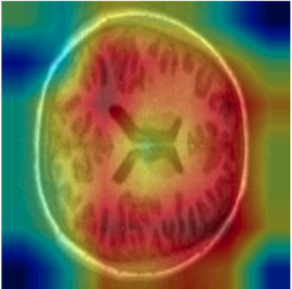}

\vspace{1mm}
{\scriptsize (a) Input slice \hspace{0.32\columnwidth} (b) GradCAM}

\caption{Slice-level evaluation: input MRI slice and corresponding GradCAM map.}
\label{fig:m1_slice_gradcam_sidebyside}
\end{figure}

The GradCAM example in Fig.~\ref{fig:m1_slice_gradcam_sidebyside} shows attention concentrated near borders and surrounding regions rather than clearly localized neuroanatomical structures. This observation motivates further examination under stricter subject-level evaluation.

\subsubsection{Subject-Level Evaluation on the Slice Dataset}

After enforcing strict subject-level separation (70/15/15), the accuracy dropped to 79.25\% on the 53-subject test set. In addition, class-wise performance became less balanced: the majority Non-demented class remained easier to identify, whereas Mild Dementia and Very Mild Dementia were more difficult to classify, resulting in a macro-F1 score of 0.435.

\begin{figure}[!t]
\centering
\begin{tabular}{cc}
\includegraphics[width=0.42\columnwidth,height=3cm]{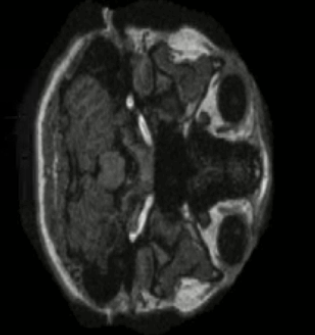} &
\includegraphics[width=0.42\columnwidth,height=3cm]{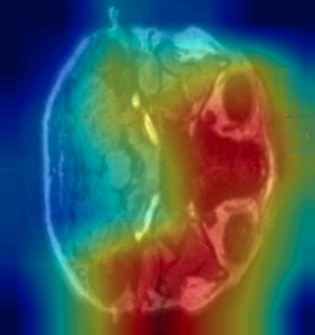} \\
{\scriptsize (a) Input slice} & {\scriptsize (b) GradCAM}
\end{tabular}

\caption{Subject-level evaluation: input MRI slice and corresponding GradCAM map.}
\label{fig:m2_slice_gradcam_sidebyside}
\end{figure}

Compared with slice level, this subject-level setting provides a more realistic estimate of slice-based generalization. However, the GradCAM example in Fig.~\ref{fig:m2_slice_gradcam_sidebyside} still suggests attention around border and background regions, indicating that the slice representation may remain sensitive to superficial cues. These supporting experiments provide context for the stronger anatomical plausibility observed in the final raw-volume multi-modal model.

\subsection{Performance and Anatomical Plausibility of the Proposed Model}

\begin{table}[!htbp]
\centering
\caption{Comparison with subject-level studies on OASIS-1. Results are not directly comparable across rows due to differences in cohort size, subject demographics, and binary grouping criteria.}
\setlength{\tabcolsep}{4pt}
\renewcommand{\arraystretch}{1.4}
\begin{tabular}{p{1.55cm} p{1.45cm} p{1.9cm} p{1.55cm} c}
\toprule
\textbf{Study} & \textbf{Dataset} & \textbf{Model} & \textbf{Input / Split} & \textbf{Acc. (\%)} \\
\midrule
Tufail et al.~\cite{tufail2020} & OASIS-1 (416) & InceptionV3, Xception & 2D slices, subject-level 5-fold CV & 63--65 \\
Yagis et al.~\cite{yagis2021}  & OASIS-1 (200) & VGG16, VGG19, ResNet-18 & 2D slices, subject-level split & 66 \\
\textbf{Ours} & \textbf{OASIS-1 (235)} & \textbf{Multi-modal 3D CNN} & \textbf{3D volumes, 5-fold subject CV} & \textbf{72.34} \\
\bottomrule
\end{tabular}
\label{tab:comparison}
\end{table}

Table~\ref{tab:comparison} places the proposed model in the context of prior subject-level studies on OASIS-1. Compared with earlier 2D slice-based approaches, the proposed multi-modal 3D CNN obtains higher accuracy under a leakage-aware evaluation setting. However, this comparison should be interpreted cautiously because prior studies differ in cohort size, preprocessing, architectures, and splitting protocols. The result suggests that combining volumetric T1 information with GM, WM, and CSF probability maps can capture useful structural patterns for Alzheimer's disease classification.

Beyond numerical performance, GradCAM visualizations show stronger attention
around medial temporal and ventricular regions, which are associated with
Alzheimer's-related structural changes~\cite{jack2018nia}. In contrast, the
supporting slice-based experiments showed stronger activation near borders
and background regions. These findings suggest that the proposed volumetric
model provides a more anatomically grounded basis for MRI-based AD
classification.
\subsection{Clinical Relevance of the Proposed Model}

In medical imaging, overly optimistic performance estimates can create a misleading impression of reliability. From this perspective, the proposed model's ROC-AUC of 0.7781 provides a more credible estimate of discriminative ability under strict subject-level evaluation. In addition, the proposed model's GradCAM maps highlight medial temporal and ventricular regions that are more consistent with known Alzheimer's-related structural changes \cite{jack2018nia}. Although this does not establish clinical readiness, it provides a more trustworthy foundation for future work on MRI-based Alzheimer's disease classification.

\section{Conclusion}
\label{sec:conclusion}

This paper presented a multi-modal 3D CNN for Alzheimer's disease
classification from raw OASIS-1 MRI volumes under strict subject-level
evaluation. By integrating structural T1 images with gray matter, white
matter, and cerebrospinal fluid probability maps, the proposed approach
provides a volumetric and anatomically grounded framework for MRI-based
analysis. The subject-level results show that the model can learn useful
structural patterns while avoiding the overly optimistic estimates that may
arise from slice-level splitting. GradCAM visualizations further suggest that
the model attends to clinically relevant brain regions, while the supporting
slice-based experiments show that data representation and evaluation protocol
can strongly influence reported performance.

At the same time, this study has several limitations. The labelled cohort is
relatively small, only two subjects belong to the moderate dementia category,
and the task was therefore formulated as binary classification. The model is
also lightweight and was not externally validated on an independent dataset.
Therefore, the findings should be interpreted as a reproducible subject-level
benchmark rather than evidence of clinical readiness. Future work should
evaluate larger and more diverse cohorts, include external validation, and
explore stronger volumetric architectures for MRI-based Alzheimer's disease
classification.

% ============================================================
\bibliographystyle{IEEEtran}
\bibliography{ref}

@misc{who2023,
  author       = {{World Health Organization}},
  title        = {Dementia},
  howpublished = {WHO Fact Sheet},
  year         = {2023},
  month        = mar,
  url          = {https://www.who.int/news-room/fact-sheets/detail/dementia}
}

@article{jack2018nia,
  author  = {Jack, C. R. and others},
  title   = {{NIA-AA} Research Framework: Toward a biological
             definition of {Alzheimer's} disease},
  journal = {Alzheimer's \& Dementia},
  volume  = {14},
  number  = {4},
  pages   = {535--562},
  year    = {2018},
  doi     = {10.1016/j.jalz.2018.02.018}
}

@article{marcus2007open,
  author  = {Marcus, D. S. and others},
  title   = {Open Access Series of Imaging Studies ({OASIS}):
             Cross-sectional {MRI} data in young, middle aged,
             nondemented, and demented older adults},
  journal = {Journal of Cognitive Neuroscience},
  volume  = {19},
  number  = {9},
  pages   = {1498--1507},
  year    = {2007},
  doi     = {10.1162/jocn.2007.19.9.1498}
}

@article{almalki2025,
  author  = {Almalki, Hassan and Khadidos, Alaa O. and
             Alhebaishi, Nawaf and Senan, Ebrahim Mohammed},
  title   = {Early detection of {Alzheimer's} disease progression
             stages using hybrid of {CNN} and transformer encoder
             models},
  journal = {Scientific Reports},
  volume  = {15},
  pages   = {16799},
  year    = {2025},
  doi     = {10.1038/s41598-025-01072-5}
}

@article{leakage2021,
  author  = {Olivetti, Emanuele and others},
  title   = {Effect of data leakage in brain {MRI} classification
             using {2D} convolutional neural networks},
  journal = {Scientific Reports},
  volume  = {11},
  pages   = {22544},
  year    = {2021},
  doi     = {10.1038/s41598-021-01681-w}
}

@misc{leakage2023,
  author        = {Rumala, Dewinda Julianensi},
  title         = {How You Split Matters: Data Leakage and Subject
                   Characteristics Studies in Longitudinal Brain
                   {MRI} Analysis},
  year          = {2023},
  eprint        = {2309.00350},
  archivePrefix = {arXiv},
  primaryClass  = {eess.IV},
  doi           = {10.48550/arXiv.2309.00350},
  note          = {arXiv preprint arXiv:2309.00350; MICCAI FAIMI 2023}
}

@article{smith2002fsl,
  author  = {Smith, Stephen M. and others},
  title   = {Advances in functional and structural {MR} image
             analysis and implementation as {FSL}},
  journal = {NeuroImage},
  volume  = {23},
  pages   = {S208--S219},
  year    = {2004},
  doi     = {10.1016/j.neuroimage.2004.07.051}
}

@article{zhang2001,
  author  = {Zhang, Yongyue and others},
  title   = {Segmentation of brain {MR} images through a hidden
             {Markov} random field model and the
             expectation-maximization algorithm},
  journal = {IEEE Transactions on Medical Imaging},
  volume  = {20},
  number  = {1},
  pages   = {45--57},
  year    = {2001},
  doi     = {10.1109/42.906424}
}

@article{tustison2010n4itk,
  author  = {Tustison, Nicholas J. and others},
  title   = {{N4ITK}: Improved {N3} bias correction},
  journal = {IEEE Transactions on Medical Imaging},
  volume  = {29},
  number  = {6},
  pages   = {1310--1320},
  year    = {2010},
  doi     = {10.1109/TMI.2010.2046908}
}

@inproceedings{selvaraju2017gradcam,
  author    = {Selvaraju, Ramprasaath R. and others},
  title     = {{Grad-CAM}: Visual explanations from deep networks
               via gradient-based localization},
  booktitle = {Proc. IEEE International Conference on Computer
               Vision (ICCV)},
  pages     = {618--626},
  year      = {2017},
  doi       = {10.1109/ICCV.2017.74}
}

@misc{tda2026,
  author        = {Ahmed, Faisal},
  title         = {Hybrid Topological and Deep Feature Fusion for
                   Accurate {MRI}-Based {Alzheimer's} Disease
                   Severity Classification},
  year          = {2026},
  eprint        = {2602.00956},
  archivePrefix = {arXiv},
  primaryClass  = {cs.CV},
  doi           = {10.48550/arXiv.2602.00956},
  url           = {https://arxiv.org/abs/2602.00956},
  note          = {arXiv preprint arXiv:2602.00956}
}

@misc{pseudocolor2025,
  author        = {Ahmed, Faisal},
  title         = {Colormap-Enhanced Vision Transformers for
                   {MRI}-Based Multiclass {Alzheimer's} Disease
                   Classification},
  year          = {2025},
  month         = dec,
  eprint        = {2512.16964},
  archivePrefix = {arXiv},
  primaryClass  = {eess.IV},
  doi           = {10.48550/arXiv.2512.16964},
  url           = {https://arxiv.org/abs/2512.16964},
  note          = {arXiv preprint arXiv:2512.16964}
}

@article{adnet2025,
  author  = {Dag, Ahmet and others},
  title   = {{ADNet}: A {CNN} Model for {Alzheimer's} Disease
             Diagnosis on {OASIS-1} Dataset},
  journal = {Karamanoglu Mehmetbey University Journal of
             Engineering and Natural Sciences},
  year    = {2025},
  month   = mar,
  doi     = {10.17780/ksujes.1534327}
}

@article{adaptivefusion2025,
  author  = {Ullah, Irfan and others},
  title   = {Hybrid Deep Learning Architecture with Adaptive
             Feature Fusion for Multi-Stage {Alzheimer's}
             Disease Classification},
  journal = {Brain Sciences},
  year    = {2025},
  volume  = {15},
  number  = {6},
  pages   = {612},
  doi     = {10.3390/brainsci15060612}
}

@inproceedings{dhinagar2021,
  author    = {Dhinagar, Nikhil J. and others},
  title     = {{3D} {CNNs} for Classification of {Alzheimer's}
               and {Parkinson's} Disease with {T1}-Weighted
               Brain {MRI}},
  booktitle = {Proc. SPIE Medical Information Processing
               and Analysis},
  volume    = {12088},
  pages     = {120880W},
  year      = {2021},
  doi       = {10.1117/12.2606297}
}

@article{kong2022,
  author  = {Kong, Zhangli and others},
  title   = {Multi-modal data {Alzheimer's} disease detection
             based on {3D} convolution},
  journal = {Biomedical Signal Processing and Control},
  volume  = {75},
  pages   = {103293},
  year    = {2022},
  doi     = {10.1016/j.bspc.2021.103293}
}

@article{peerj2025,
  author  = {Zhang, Xin and others},
  title   = {Classification of the stages of {Alzheimer's}
             disease based on three-dimensional lightweight
             neural networks},
  journal = {PeerJ Computer Science},
  volume  = {11},
  pages   = {e2897},
  year    = {2025},
  doi     = {10.7717/peerj-cs.2897}
}

@article{baltrusaitis2019,
  author  = {Baltru\v{s}aitis, Tadas and Ahuja, Chaitanya
             and Morency, Louis-Philippe},
  title   = {Multimodal Machine Learning: A Survey and Taxonomy},
  journal = {IEEE Transactions on Pattern Analysis and
             Machine Intelligence},
  volume  = {41},
  number  = {2},
  pages   = {423--443},
  year    = {2019},
  doi     = {10.1109/TPAMI.2018.2798607}
}

@article{multimodal_vit2024,
  author  = {Yousaf, Taimoor and others},
  title   = {A multimodal vision transformer for interpretable
             fusion of functional and structural neuroimaging
             data},
  journal = {Human Brain Mapping},
  volume  = {45},
  number  = {17},
  year    = {2024},
  doi     = {10.1002/hbm.26800}
}

@inproceedings{lin2014gap,
  author    = {Lin, Min and Chen, Qiang and Yan, Shuicheng},
  title     = {Network In Network},
  booktitle = {Proc. International Conference on Learning
               Representations (ICLR)},
  year      = {2014}
}

@article{srivastava2014dropout,
  author  = {Srivastava, Nitish and others},
  title   = {Dropout: A Simple Way to Prevent Neural Networks
             from Overfitting},
  journal = {Journal of Machine Learning Research},
  volume  = {15},
  pages   = {1929--1958},
  year    = {2014}
}

@article{tufail2020,
  author  = {Tufail, Ahsan Bin and others},
  title   = {Binary Classification of {Alzheimer's} Disease Using {sMRI} Imaging
             Modality and Deep Learning},
  journal = {Journal of Digital Imaging},
  year    = {2020},
  note    = {PMC7573078}
}

@article{yagis2021,
  author  = {Yagis, Ekin and others},
  title   = {Effect of Data Leakage in Brain {MRI} Classification Using {2D}
             Convolutional Neural Networks},
  journal = {Scientific Reports},
  year    = {2021},
  volume  = {11},
  note    = {PMC8604922}
}

\end{document}